\documentclass[journal=jacsat,manuscript=article]{achemso}

\usepackage{chemformula} % Formula subscripts using \ch{}
\usepackage[T1]{fontenc} % Use modern font encodings

\usepackage{graphicx}
\usepackage{multirow}
\usepackage{hyperref}
\hypersetup{
    colorlinks=true,
    linkcolor=blue,
    filecolor=magenta, 
    urlcolor=blue,
    breaklinks=true
}
\usepackage{multirow}
\usepackage{threeparttable}
\usepackage{booktabs}
\usepackage{float}
\usepackage{color}
\usepackage{soul}

\author{Zun Wang$^{1,2}$}
\author{Chong Wang$^3$}
\email{ch-wang@outlook.com}
\author{Sibo Zhao$^1$}
\author{Yong Xu$^{1,2,5,6}$}
\author{Shaogang Hao$^2$}
\email{shaoganghao@tencent.com}
\author{Chang Yu Hsieh$^2$}
\author{Bing-Lin Gu$^{1,4}$}
\author{Wenhui Duan$^{1,2,4,5}$}
\email{duanw@tsinghua.edu.cn}
\affiliation{$^1$State Key Laboratory of Low Dimensional Quantum Physics and Department of Physics, Tsinghua University, Beijing, 100084, China\\
$^2$Tencent Quantum Lab, Tencent, Shenzhen, Guangdong 518057, China\\
$^3$Department of Physics, Carnegie Mellon University, Pittsburgh, Pennsylvania 15213, USA\\
$^4$Institute for Advanced Study, Tsinghua University, Beijing 100084, China\\
$^5$Frontier Science Center for Quantum Information, Beijing 100084, China\\
$^6$RIKEN Center for Emergent Matter Science (CEMS), Wako, Saitama 351-0198, Japan}

\title[An \textsf{achemso} demo]{Heterogeneous relational message passing networks for molecular dynamics simulations}

\begin{document}

% \begin{tocentry}

% Some journals require a graphical entry for the Table of Contents.
% This should be laid out ``print ready'' so that the sizing of the

% Inside the \texttt{tocentry} environment, the font used is Helvetica
% 8\,pt, as required by \emph{Journal of the American Chemical

% The surrounding frame is 9\,cm by 3.5\,cm, which is the maximum
% permitted for  \emph{Journal of the American Chemical Society}
% graphical table of content entries. The box will not resize if the
% content is too big: instead it will overflow the edge of the box.

% This box and the associated title will always be printed on a
% separate page at the end of the document.

% \end{tocentry}

\begin{abstract}
  With many frameworks based on message passing neural networks proposed to predict molecular and bulk properties, machine learning methods have tremendously shifted the paradigms of computational sciences underpinning physics, material science, chemistry, and biology. While existing machine learning models have yielded superior performances in many occasions, most of them model and process molecular systems in terms of homogeneous graph, which severely limits the expressive power for representing diverse interactions. In practice, graph data with multiple node and edge types is ubiquitous and more appropriate for molecular systems. Thus, we propose the heterogeneous relational message passing network (HermNet), an end-to-end heterogeneous graph neural networks, to efficiently express multiple interactions in a single model with {\it ab initio} accuracy. HermNet performs impressively against many top-performing models on both molecular and extended systems. Specifically, HermNet outperforms other tested models in nearly 75\%, 83\% and 94\% of tasks on MD17, QM9 and extended systems datasets, respectively. Finally, we elucidate how the design of HermNet is compatible with quantum mechanics from the perspective of the density functional theory. Besides, HermNet is a universal framework, whose sub-networks could be replaced by other advanced models.
\end{abstract}

\section{Introduction}
In the realm of physics, chemistry, material science, and biology, multi-scale modeling~\cite{weinan2003multiscale, horstemeyer2009multiscale} helps us understand the properties of materials in multiple scales of time and space. Molecular dynamics (MD) simulation is an essential tool for modeling dynamical evolution of a many-body system. The trajectories of interacting particles are determined by solving Newton's equations of motion involving complex interatomic potentials. There are two mainstream approaches for performing MD simulations, i.e., classical molecular dynamics~\cite{alder1959studies} and {\it ab initio} molecular dynamics (AIMD)~\cite{car1985unified}. The potential energy surface in classical molecular dynamics is given by parameterized force fields of a presumed functional form, which facilitates large-scale calculations but possesses poor transferability across tasks. On the other hand, AIMD computes the total energy of a system using quantum mechanics methods, such as the density functional theory (DFT)~\cite{kohn1965self}, that guarantees the applicability and the accuracy under a wide variety of conditions. However, due to the cost of rigorously treating the electronic degrees of freedom, AIMD modeling is currently limited to physical and chemical systems of modest scales. With the rapid development of technology for chemical and material synthesis, the need to construct force fields for large-scale calculations with accuracy comparable to that of the first-principles methods has become ever more urgent.

One recent development to address the above issue is to use machine learning methods~\cite{jordan2015machine, lecun2015deep} to  facilitate MD simulations. The most important tool in machine learning is neural networks. The first framework of neural networks for MD simulations is proposed by Behler and Parrinello~\cite{behler2007generalized}, which is based on fully connected neural networks. Considerable success has been achieved along this route. Especially, Deep potential (DeePMD)~\cite{zhang2018deep, zhang2018end} has been developed as a comprehensive software suite and has been used in simulations of crystal nucleation~\cite{bonati2018silicon, niu2020ab} and construction of phase diagram~\cite{zhang2021phase}. Traditional neural networks, for example, fully connected neural networks and convolutional neural networks, are most useful when the input data are Euclidean. However, atoms are intrinsically indistinguishable and cannot be ordered. As a result, heavy data preprocessing have to be performed in the above-mentioned frameworks. To alleviate such data preprocessing burden, graph neural networks (GNN)~\cite{zhou2020graph, wu2020comprehensive} are introduced. The power of graph formalism lies in its focus on relationships among entities (or nodes) rather than the properties of individual nodes. In particular, message passing neural networks (MPNN)~\cite{gilmer2017neural} summarized the recapitulative formula for GNN in the spatial domain. With atoms represented as nodes and interactions or bonds between them represented as edges in a graph, molecules or crystals can be transformed to molecular graphs or crystal graphs naturally. GNN-based frameworks for MD simulations, including DTNN~\cite{schutt2017quantum}, SchNet~\cite{schutt2017schnet, schutt2018schnet}, DimeNet~\cite{klicpera2020directional, klicpera2020fast}, PAINN~\cite{schutt2021equivariant}, and MDGNN~\cite{wang2021symmetry}, have accurately predicted the potential surface of small molecules and crystals. Current GNN-based MD simulations mostly use homograph, where the message passing network is the same regardless of the types of the atoms. On the other hand, it is now a common practice to use the hybrid pair style in MD simulations, which utilizes different force fields for atom pairs of different types. The hybrid pair style is very useful for complex material systems, such as polymers on metal surface, polymers with nano-particles and solid-solid interface between two different materials. This motivates us to explore the possibility of improved performance by using heterogeneous graph in GNN-based MD simulations.

In this work, we propose a framework to model diverse interactions in a single molecular dynamics simulations, termed heterogeneous molecular dynamics networks (HermNet). The model shares a similar idea of hybrid pair style in Large-scale Atomic/Molecular Massively Parallel Simulator software (LAMMPS)~\cite{plimpton1995fast}. HermNet splits the molecular or crystal graph into several subgraphs and use different message passing networks for different subgraphs. Within each subgraph, we choose a modified version of polarizable atom interaction neural network (PAINN)~\cite{schutt2021equivariant} as the subnetwork. Experiments on molecular and extended systems were performed and the results were satisfactory. HermNet provides a general method to design heterogeneous GNN for MD simulations.

\section{Preliminary}
In the graph theory~\cite{bondy1976graph}, a graph is a data structure composed of sets of vertices and edges. Graphs could be classified either as undirected graphs or digraphs by whether there is an explicit designation of edges' orientations. From the standpoint that an undirected edge graph can be interpreted as a bidirectional link between the pair of nodes, undirected graphs are made up of digraphs.

Graphs can be further classified either as homogeneous or heterogeneous, according to the types of nodes and edges. A homogeneous graph is a special case of heterogeneous graph. MPNN~\cite{gilmer2017neural}, which is a universal spatial-domain-based graph neural network framework, was proposed for homogeneous graphs. With $h_v$ and $e_{vw}$ denoting, respectively, node features and edge features in a graph, MPNN is summarized as
\begin{align}
    m_v^{t+1} &= \sum_{w\in\mathcal{N}(v)} M_t(h_v^t, h_w^t, e_{vw}^t),\\
    h_v^{t+1} &= U_t(h_v^t, m_v^{t+1}),
\end{align}
where the forward propagation is decomposed into two phases, a message passing phase and a readout phase. $M_t$ and $U_t$ are a message function and a update function, respectively. The hidden states $h_w$ of all the neighbors $\mathcal{N}(v)$ of vertex $v$ will be aggregated and then be used to update hidden states of vertex $v$ in the next step. A heterogeneous graph supports sophisticated multi-type relations and inherently enables richer semantic relations. Relational graph convolutional network (R-GCN)~\cite{schlichtkrull2018modeling} is an extension of MPNN. $\mathcal{G}=(\mathcal{V}, \mathcal{E}, \mathcal{R})$ denotes a heterogeneous graph with nodes (entities) $v_i\in\mathcal{V}$ and labeled edges (relations) $(v_i, r, v_j)\in\mathcal{E}$, where $r\in\mathcal{R}$ is a relation type, that covers both canonical directional and inverse directional relations. A generalized forward process of an entity $v_i$ in a relational graph takes the form
\begin{equation}\label{eq:rgcn-update}
    h_i^{(l+1)} = \sum_{r\in\mathcal{R}}U_r^{(l)}\left(h_i^{(l)}, \sum_{j\in\mathcal{N}_i^r} M_r^{(l)}(h_i^{(l)}, h_j^{(l)}, e_{ij}^{(l)})\right),
\end{equation}
where $\mathcal{N}_i^r$ denotes the set of neighbor indices of vertex $i$ of relation $r$. Eq.~\ref{eq:rgcn-update} implies that a heterogeneous graph can be decomposed into several homogeneous graphs of distinct relations $\mathcal{R}$. Typically, each homogeneous graph is a directed graph. In other words, an R-GCN layer is made up of multiple MPNN layers, each of which is associated with a homogeneous graph of relation $r$.

\begin{figure}
    \centering
    \includegraphics[width=0.9\linewidth]{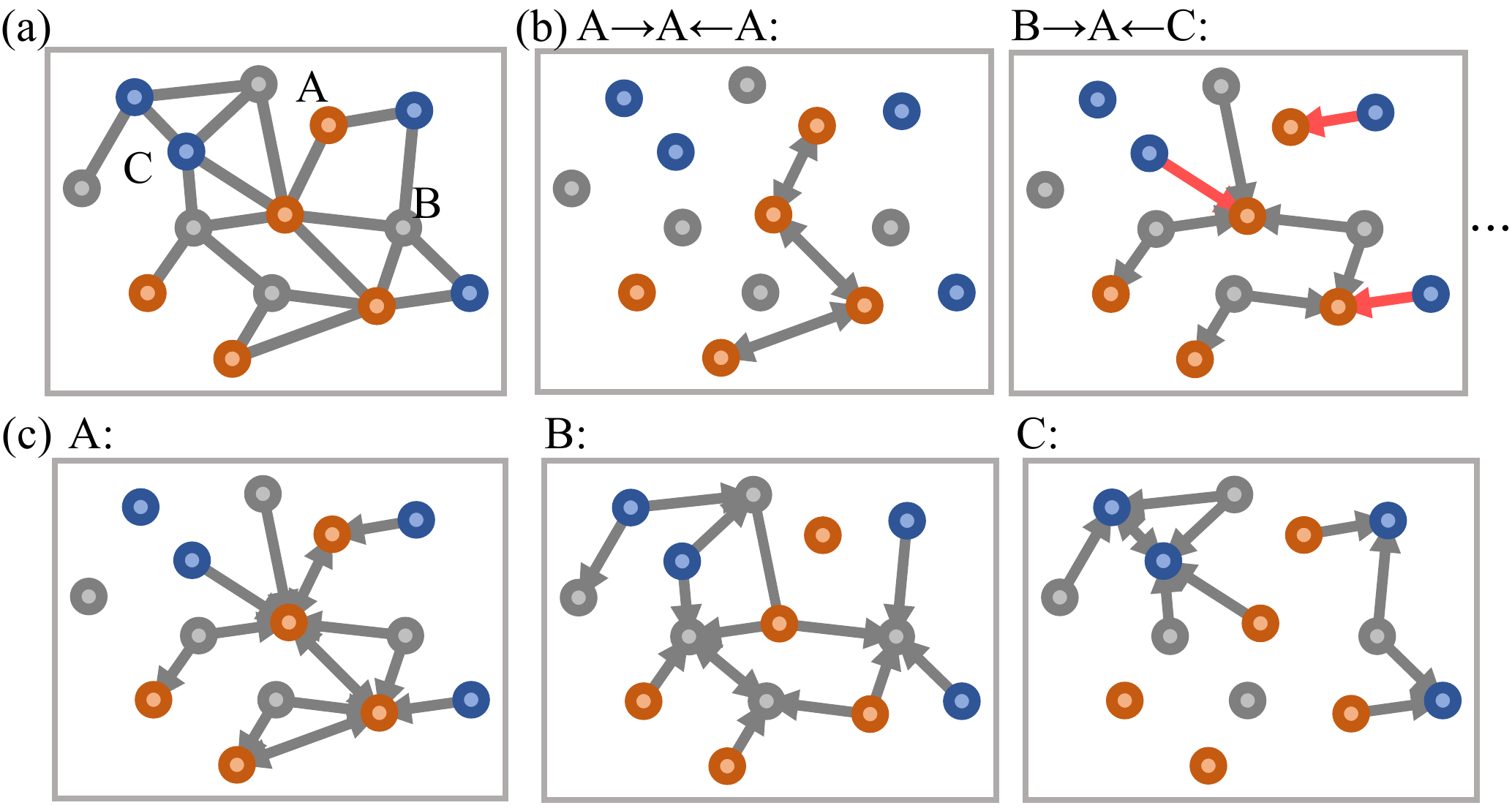}
    \caption{A schematic diagram that demonstrates how to extract subgraphs from a original heterogeneous graph. (a) The original graph constructed via a certain method with multiple node types, specifically, three types A, B, and C here. (b) Subgraphs extracted from the original graph according to triadic relations. The number of subgraphs is of the order of $N_e^3$, where $N_e$ is the number of node types. (c) Subgraphs extracted from the original graph according to type of central nodes. In this case, the number of subgraphs linearly increases with respect to $N_e$.}
    \label{fig:1}
\end{figure}

\begin{figure*}
    \centering
    \includegraphics[width=0.8\linewidth]{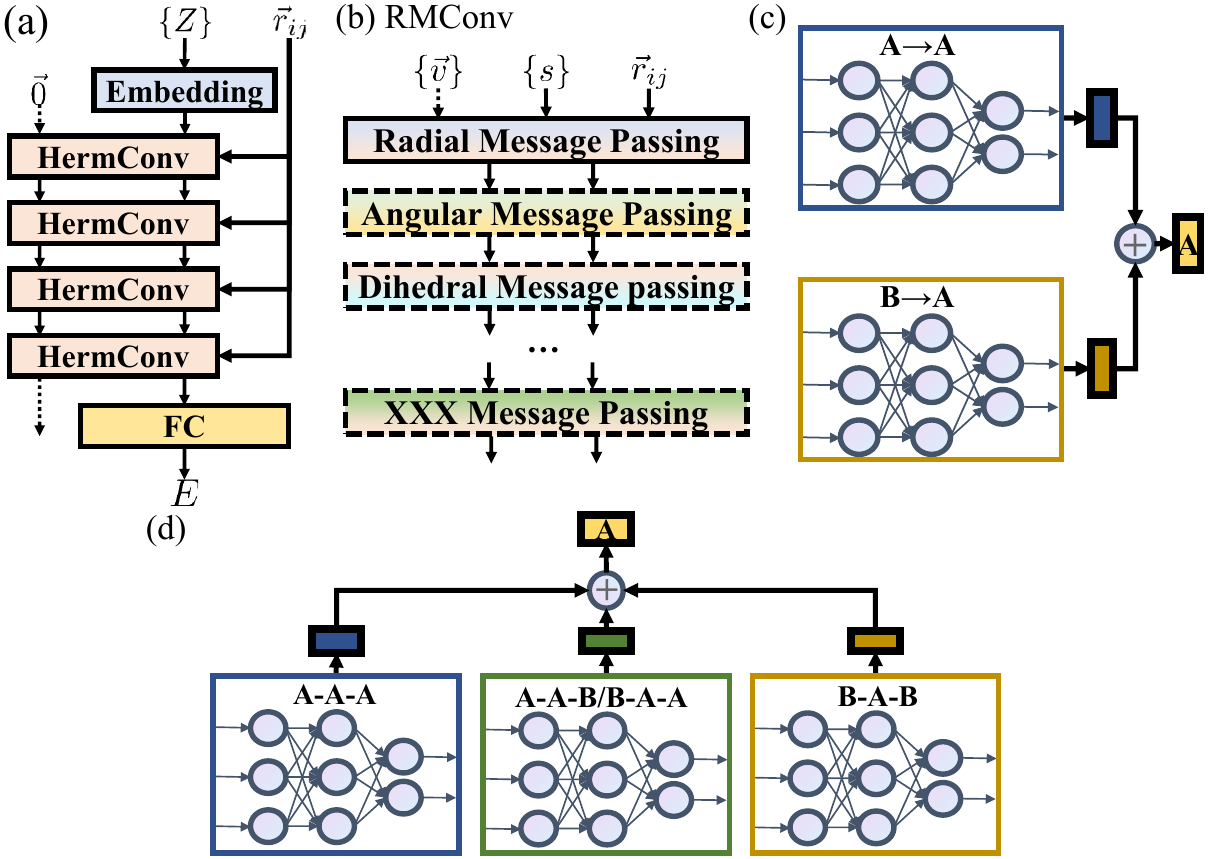}
    \caption{The schematic of the working principle of the entire architecture. (a) The entire architecture diagram of HermNet where $\{Z\}$ is the set of atomic numbers, which will be passed through an embedding layer. Initial vectorial node features are all-zero vectors of fixed dimension. This layer is expected to receive a scalar node feature $\{s\}$, a vectorial node feature $\{\vec{v}\}$ and a vectorial edge feature, i.e. relative position vector $\vec{r}_{ij}$, and then output an updated scalar and vectorial node feature as the inputs of the next layer. The final scalar node features will be passed to a global pooling layer as feature of the graph. With the graph-level feature passing to a sequence of fully-connected layers, the target to predict is achieved. (b) Sub-network for processing related subgraphs, i.e. homogeneous digraphs. The layer is composed of message passing layers hierarchically, such as radial message passing layer for two-body interactions, angular message passing layer for three-body interactions, and so on. Related message passing layers will be truncated according to the level of interactions to be modelled. The features or/and message passing layers with dotted line should be introduced in accordance with requirements. Several sub-networks which model different relations compose a single heterogeneous relational message passing layer. When the interactions are truncated to two-body interactions, the entire framework is termed HPNet. (c) Sub-network in HPNet for A-type when the system contains only two kinds of elements, specifically, A- and B-type. (d) The hidden states of A-type vertex derive from a sub-network that is truncated to three-body interactions for corresponding relations. The colors of the networks for different three-body interactions represent the parameters. If these colors are the same, which means the parameters are shared in all these three networks, the HermNet is termed HVNet. If not, then the HermNet is termed HTNet.}
    \label{fig:2}
\end{figure*}

\section{Architecture}
Diverse forms of force fields are manifestly responsible for the intricate interactions, especially in systems with multiple elements. Graph neural networks for homogeneous graphs model interactions of different atomic pairs with shared parameters, which limits the expressive power for neural-network-based force fields. For example, as shown in Fig.~\ref{fig:1} (a), there are three kinds of particles, i.e. A-, B- and C-type atoms. The graph is constructed via linking central nodes with their adjacent nodes within a cutoff radius. In a classical molecular dynamics simulation for this system, six different force fields can be allocated for A-A pairs, A-B pairs, A-C pairs, etc., provided only two body interactions are considered. 
If a homogeneous graph neural network is employed to model different interactions by fitting a single function, it is expected to generate a mean force field. On the other hand, equipped with multiple types of nodes and edges, a heterogeneous graph neural network is a natural choice to model these interactions with a more detailed resolution.

\begin{figure*}
    \centering
    \includegraphics[width=0.9\linewidth]{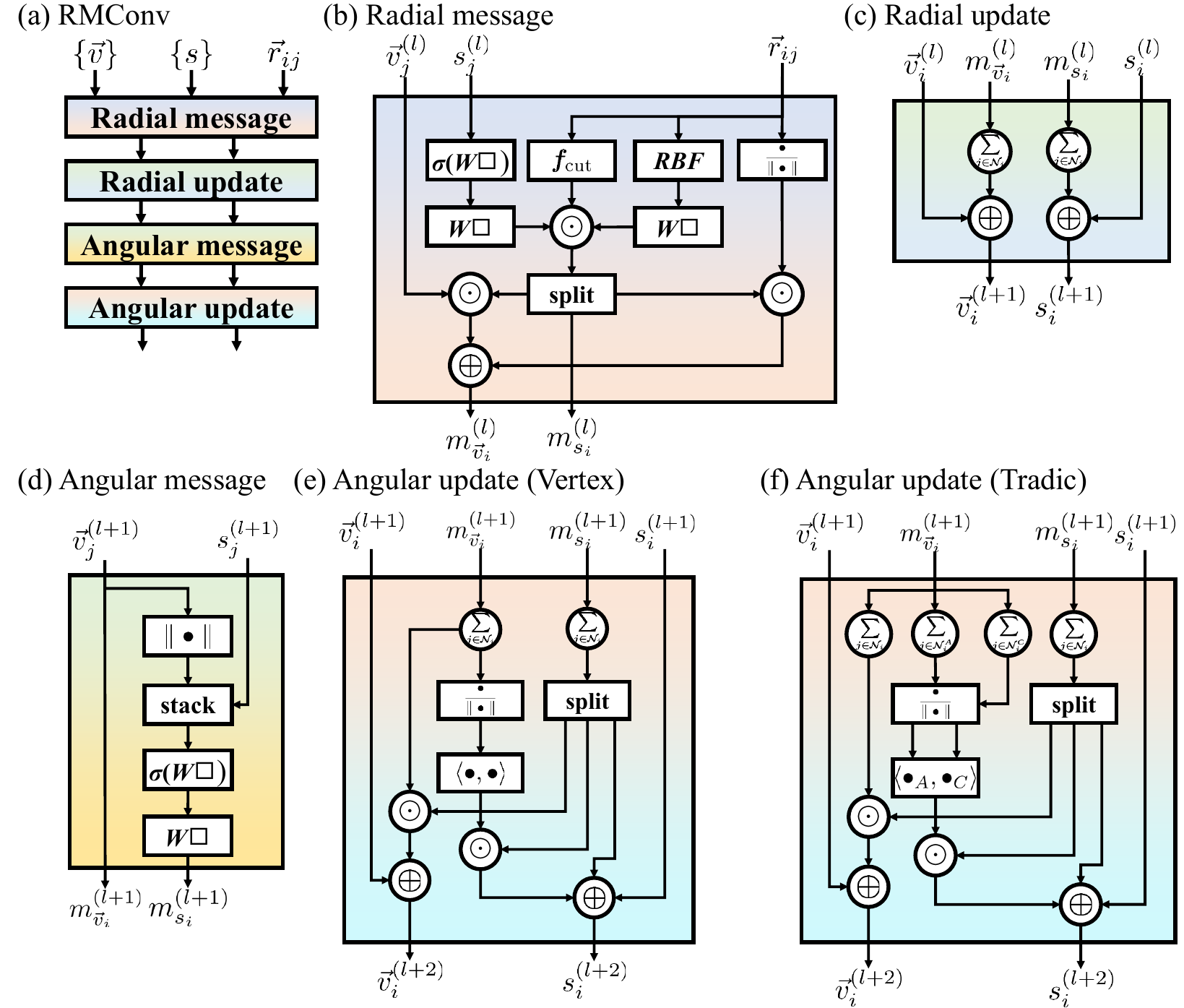}
    \caption{The overview of the sub-networks in HVNet and HTNet. (a) is the architecture of the sub-network, termed relational message passing convolutional (RMConv) layer. Such a RMConv layer is a simplified and modified PAINN~\cite{schutt2021equivariant} invoked for a specific type of interaction, and is constituted by (b) radial message layer, (c) radial update layer, (d) angular message layer, and (e) or (f) angular update layer. $\{s\}$ is the set of scalar node features and initially set as the atomic numbers, which will be passed through an embedding layer. Initial vectorial node feature $\vec{v}^{(0)}$ is an all-zero vector with a fixed dimension. $\sin\left(\frac {n\pi} {r_\text{cut}}\parallel \vec{r}_{ij}\parallel\right) / \parallel \vec{r}_{ij}\parallel$ with $1 \leq n \leq 30$ are selected as radial basis functions (RBF)~\cite{klicpera2020directional} and a cosine cutoff $f_\text{cut}$~\cite{behler2011atom} is also adopted in the filter. The original message layer in PAINN~\cite{schutt2021equivariant} (i.e., a MPNN layer) is decomposed into (b) and (c) (i.e., radial message layer and radial update layer). A modified and simplified update layer is decomposed into (d) and (e) for HVNet or (f) for HTNet. These layers model three-body interactions via expressing angular information explicitly.}
    \label{fig:3}
\end{figure*}

As shown in the following, we develop a universal framework, HermNet, to model diverse many-body interactions simultaneously via extracting appropriate subgraphs, which are subsequently processed by heterogeneous graph neural networks. The overview of the entire architecture diagram of HermNet is displayed in Fig.~\ref{fig:2} (a), which takes atomic numbers $Z$ (and a vector of zeros) as the node's scalar features (and node's vectorial features). HermNet is composed of several message passing layers, termed HermConv layers (Fig.~\ref{fig:2} (b)), which model interactions hierarchically. We introduce three variants of HermNets: heterogeneous pair networks (HPNet), heterogeneous triadic networks (HTNet), and heterogeneous vertex networks (HVNet). A HPNet layer for central nodes of A-type is displayed in Fig.~\ref{fig:2} (c), where all the sub-networks with A-type destination contribute to the local environment of A-type node. A HermNet layer for HVNet and HTNet is displayed in Fig.~\ref{fig:2} (d). If the parameters of its sub-networks (RMConv, see Fig.~\ref{fig:3}) are shared for the same kinds of central nodes, this HermNet framework is referred to as HVNet. When the parameters are not shared, this HermNet framework is a HTNet. We only test and report HVNet's performance in the following sections, as the other two models (HPNet and HTNet) have high complexity and will take more training time and data points for a proper assessment.

Most machine learning frameworks for MD simulation only take into account the interatomic distances in feature engineering, ignoring the bond angle information, which is an important characteristic of both molecules and crystals. In principle bond angle can be deduced from interatomic distances. However, it is advantageous to explicitly include bond angle information in feature engineering to achieve better performance. Directional message passing networks (DimeNet)~\cite{klicpera2020directional, klicpera2020fast} innovatively introduced three-body interactions explicitly by combining radial and angular information from the edges of the original graph and the corresponding line graph, respectively. PAINN~\cite{schutt2021equivariant} is a rotationally equivariant MPNN framework and the complexity of calculating angular information was reduced. In this work, we incorporate angular information by choosing PAINN as the sub-network in HermNet. This specific message-passing setup can be directly implemented in HVNet, while slight modifications are required in HTNet to distinguish the type of source nodes. We note that HPNet cannot incorporate all angular information explicitly. For example, the bond angle A$\rightarrow$B$\leftarrow$B is lost in HPNet because A$\rightarrow$B and B$\leftarrow$B are processed by different sub-networks.

As discussed above, a heterogeneous graph could be decomposed into several homogeneous subgraphs. To describe the method of extracting these subgraphs, we use $\mathcal{G}$, $\hat{\mathcal{Q}}_s$, and $\hat{\mathcal{Q}}_d$ to denote the input heterogeneous graph, the operator that returns the subgraphs with specific source nodes, and the operator that returns the subgraphs with specific destination nodes, respectively. As indicated in Fig.~\ref{fig:1} (b) and (c), the directed subgraphs for HVNet could be extracted via selecting inbound edges of a given A-type destination node, i.e. $\hat{\mathcal{Q}}_d^A\mathcal{G}$, while those for HTNet are extracted via selecting inbound edges of a given B-type destination node firstly and then choosing out-bound edges of its A-type and C-type source nodes simultaneously, i.e. $\hat{\mathcal{Q}}_s^{A\cup C}\hat{\mathcal{Q}}_d^B\mathcal{G}$ for triadic relation A$\rightarrow$B$\leftarrow$C. We note that if the two destination nodes are extracted sequentially for HTNet, the result is generally an empty graph.

\section{Results}
In the following, we report the testing of HVNet against other prior frameworks on three well-established benchmark datasets. As detailed below, HVNet convincingly outperforms most of the prior methods.

\subsection{Benchmarks on QM9 dataset}

\begin{table*}
\centering
  {\caption{Comparison of the MAEs between several benchmarked models and HVNet trained on MD17 dataset using 1000 training samples (energies in meV and forces in meV/\AA). The values in bold represent outperformance on the same task.}
  \label{tab:1}
  \vspace{2mm}
  \begin{threeparttable}
  \begin{tabular}{rrrrrrrrrrr}
      \hline\hline
      ~ & \multicolumn{1}{c|}{~} & \multicolumn{1}{c}{\textbf{SchNet\textsuperscript{\emph{a}}}} & ~ & \multicolumn{1}{c}{\textbf{DimeNet\textsuperscript{\emph{b}}}} & ~ & \multicolumn{1}{c}{\textbf{PAINN\textsuperscript{\emph{c}}}} & ~ &\multicolumn{1}{c}{\textbf{HVNet}}\\
      \midrule[1pt]
      \multicolumn{1}{c|}{\multirow{2}{*}{Benzene}} & \multicolumn{1}{c|}{energy} & 3.44 & ~ & 3.354 & ~ & - & ~ & \textbf{0.319}\\
      \cline{2-9}
      \multicolumn{1}{c|}{~} & \multicolumn{1}{c|}{force} & 13.33 & ~ & 8.041 & ~ & - & ~ & \textbf{1.223}\\
      \hline
      \multicolumn{1}{c|}{\multirow{2}{*}{Toluene}} & \multicolumn{1}{c|}{energy} & 5.16 & ~ & 4.386 & ~ & 4.171 & ~ & \textbf{1.967}\\
      \cline{2-9}
      \multicolumn{1}{c|}{~} & \multicolumn{1}{c|}{force} & 24.51 & ~ & 9.288 & ~ & \textbf{4.386} & ~ & 5.268\\ 
      \hline
      \multicolumn{1}{c|}{\multirow{2}{*}{Malonaldehyde}} & \multicolumn{1}{c|}{energy} & 5.59 & ~ & 4.472 & ~ & 3.913 & ~ & \textbf{1.422}\\
      \cline{2-9}
      \multicolumn{1}{c|}{~} & \multicolumn{1}{c|}{force} & 28.38 & ~ & 16.469 & ~ & 13.717 & ~ & \textbf{6.215}\\
      \hline
      \multicolumn{1}{c|}{\multirow{2}{*}{Salicylic acid}} & \multicolumn{1}{c|}{energy} & 8.60 & ~ & 5.762 & ~ & 4.902 & ~ & \textbf{4.128}\\
      \cline{2-9}
      \multicolumn{1}{c|}{~} & \multicolumn{1}{c|}{force} & 36.55 & ~ & 16.082 & ~ & \textbf{8.987} & ~ & 10.887\\
      \hline
      \multicolumn{1}{c|}{\multirow{2}{*}{Aspirin}} & \multicolumn{1}{c|}{energy} & 15.91 & ~ & 8.772 & ~ & \textbf{6.837} & ~ & 9.935\\
      \cline{2-9}
      \multicolumn{1}{c|}{~} & \multicolumn{1}{c|}{force} & 58.05 & ~ & 21.457 & ~ & 15.953 & ~ & \textbf{11.734}\\
      \hline
      \multicolumn{1}{c|}{\multirow{2}{*}{Ethanol}} & \multicolumn{1}{c|}{energy} & 3.44 & ~ & 2.752 & ~ & 2.709 & ~ & \textbf{1.258}\\
      \cline{2-9}
      \multicolumn{1}{c|}{~} & \multicolumn{1}{c|}{force} & 16.77 & ~ & 9.89 & ~ & 9.89 & ~ & \textbf{6.102}\\
      \hline
      \multicolumn{1}{c|}{\multirow{2}{*}{Uracil}} & \multicolumn{1}{c|}{energy} & 6.02 & ~ & 4.945 & ~ & 4.472 & ~ & \textbf{1.660}\\
      \cline{2-9}
      \multicolumn{1}{c|}{~} & \multicolumn{1}{c|}{force} & 24.08 & ~ & 12.943 & ~ & 6.02 & ~ & \textbf{3.999}\\
      \hline
      \multicolumn{1}{c|}{\multirow{2}{*}{Naphtalene}} & \multicolumn{1}{c|}{energy} & 6.88 & ~ & 5.246 & ~ & 5.031 & ~ & \textbf{2.728}\\
      \cline{2-9}
      \multicolumn{1}{c|}{~} & \multicolumn{1}{c|}{force} & 24.94 & ~ & 9.245 & ~ & \textbf{3.569} & ~ & 4.469\\
      \hline\hline
  \end{tabular}
  \textsuperscript{\emph{a}} Ref.~\citenum{schutt2017schnet, schutt2018schnet}
  \textsuperscript{\emph{b}} Ref.~\citenum{klicpera2020directional}
  \textsuperscript{\emph{c}} Ref.~\citenum{schutt2021equivariant}
  \end{threeparttable}}
\end{table*}

The QM9 dataset~\cite{ruddigkeit2012enumeration, ramakrishnan2014quantum} consists of computed geometric, energetic, electronic, and thermodynamic properties for 134k stable small organic molecules made up of carbon, hydrogen, oxygen, nitrogen, and fluorine. All properties were calculated at the B3LYP/6-31G (2df, p) level of quantum chemistry. This dataset provides quantum chemical insights for the relevant chemical space of small organic molecules, and has been widely adopted as the benchmark to calibrate, analyze and evaluate new methods in this field. HVNet was trained on 110k molecules and validated on another 10k molecules. The properties of the 134k molecules include dipole moment ($\mu$), isotropic polarizability ($\alpha$), energy of the highest occupied molecular orbital ($\varepsilon_\text{HOMO}$), energy of the lowest unoccupied molecular orbital ($\varepsilon_\text{LUMO}$), band gap ($\Delta \varepsilon$), electronic spatial extent ($R^2$), zero point vibrational energy ($ZPVE$), internal energy at 0 K ($U_0$), internal energy at 298.15 K ($U$), enthalpy at 298.15 K ($H$), free energy at 298.15 K ($G$), and heat capacity at 298.15 K ($c_v$). It must be emphasized that HVNets were trained with atomization energies rather than the original internal energies, enthalpy energy, and free energy, i.e., the original energies subtracting the atomic reference energies, which is the protocol advocated in the DimeNet work of Klicpera {\it et al.} ~\cite{klicpera2020directional}. These adjusted values are more reasonable because absolute energies are generally meaningless and relative energies essentially convey all physical implications. Table~\ref{tab:2} reports the MAEs of HVNet for 12 tasks in the QM9 dataset with comparison to other eight models. HVNet outperforms all baselines on 10 out of 12 tasks. For the other 2 tasks, $R^2$ and ZPVE, the MAEs of HVNet are on par with some of the baselines. Details of additional settings and the definition of the physical quantities with respect to the models and datasets are provided in the Supplemental Material (SM)~\cite{SI}.

\subsection{Benchmarks on MD17 dataset}
The MD17 dataset~\cite{chmiela2017machine, schutt2017quantum, chmiela2018towards} provides non-equilibrium structures sampled (at a time resolution of 0.5 fs) from AIMD trajectories for eight small molecules with a background temperature of 500 K. The potential energy and force labels are computed with PBE+vdW-TS method. Anders {\it et al.} found that the energies in original MD17 dataset are contaminated with substantial numerical noises and published a revised version of the MD17~\cite{christensen2020role} dataset. Distinct HVNet models were trained on this revised dataset, and an a 1000-frame training set and a 1000-frame validation set are randomly selected. The learning rate was initially set at $3\times 10^{-4}$ and adaptively reduced when the loss on the validation set reached a plateaus. The truncated radius was set at 5 \AA~for the construction of molecular graphs. Additional details can be found in the Supplemental Material (SM)~\cite{SI}. Table~\ref{tab:1} presents the comparisons of mean absolute errors (MAEs) of three benchmarked models and HVNet. HVNet outperforms other models with a comfortable margin on three quarters of the predictive tasks, and its results of the remaining tasks are comparable to the best results among all four frameworks. We also attempted to train a HTNet on the MD17 dataset; however, the parameter space of the HTNet is simply too large, and obvious overfitting was immediately observed after just several training epochs. Nevertheless, we believe that HTNet has the capability to express the force fields once enough data points are provided.

\begin{table*}
% \begin{sidewaystable}[thp]
\centering
  {\caption{Comparison of the MAEs between several benchmark models and HVNet trained on QM9 dataset. The values in bold represent outperformance on the same task.}
  \label{tab:2}
  \resizebox{\textwidth}{28mm}{
  \vspace{2mm}
  \begin{threeparttable}
  \begin{tabular}{c|c|rrrrrrrrr}
      \hline\hline
      ~ & units & \textbf{SchNet\textsuperscript{\emph{a}}} &\textbf{DimeNet\textsuperscript{\emph{b}}}&\textbf{DimeNet++\textsuperscript{\emph{c}}}&\textbf{Cormorant\textsuperscript{\emph{d}}}&\textbf{HMGNN\textsuperscript{\emph{e}}}&\textbf{MXMNet\textsuperscript{\emph{f}}}&\textbf{PAINN\textsuperscript{\emph{g}}}&\textbf{DeepMoleNet\textsuperscript{\emph{h}}} &\textbf{HVNet}\\
      \midrule[1pt]
      $\mu$ & $D$ & 0.033 & 0.0286 & 0.0297 & 0.038 & 0.0272 & 0.0255 & 0.012 & 0.0178 & \textbf{0.00352}\\
      $\alpha$ & $a_0^3$ & 0.235 & 0.0469 & 0.0435 & 0.085 & 0.0561 & 0.0447 & 0.045 & 0.0475 & \textbf{0.0327}\\
      $\epsilon_{\text{HOMO}}$ & meV & 41 & 27.8 & 24.6 & 34 & 24.78 & 22.8 & 27.6 & 21.9 & \textbf{1.385}\\
      $\epsilon_{\text{LUMO}}$ & meV & 34 & 19.7 & 19.5 & 38 & 20.61 & 18.9 & 20.4 & 18.5 & \textbf{3.265}\\ 
      $\Delta \epsilon$ & meV & 63 & 34.8 & 32.6 & 38 & 33.31 & 30.6 & 45.7 & 32.1 & \textbf{3.732}\\
      $R^2$ & $a_0^2$ & 0.073 & 0.331 & 0.331 & 0.961 & 0.416 & 0.088 & \textbf{0.066} & 0.115 & 0.369\\
      $ZPVE$ & meV & 1.70 & 1.29 & 1.21 & 2.03 & 1.18 & \textbf{1.15} & 1.28 & 1.22 & 1.949\\
      $U_0$ & meV & 14 & 8.02 & 6.32 & 22 & 5.92 & 5.9 & 5.85 & 6.1 & \textbf{4.512}\\
      $U$ & meV & 19 & 7.89 & 6.28 & 21 & 6.85 & 5.94 & 5.83 & 6.1 & \textbf{5.445}\\
      $H$ & meV & 14 & 8.11 & 6.53 & 21 & 6.08 & 6.09 & 5.98 & 6.1 & \textbf{5.098}\\
      $G$ & meV & 14 & 8.98 & 7.56 & 20 & 7.61 & 7.17 & 7.35 & 7.1 & \textbf{6.729}\\
      $c_v$ & $\frac{\text{cal}}{\text{mol K}}$ & 0.033 & 0.0249 & 0.0230 & 0.026 & 0.0233 & 0.0224 & 0.024 & 0.0241 & \textbf{0.01964}\\
      \hline\hline
  \end{tabular}
  \textsuperscript{\emph{a}} Ref.~\citenum{schutt2017schnet, schutt2018schnet}
  \textsuperscript{\emph{b}} Ref.~\citenum{klicpera2020directional}
  \textsuperscript{\emph{c}} Ref.~\citenum{klicpera2020fast}
  \textsuperscript{\emph{d}} Ref.~\citenum{anderson2019cormorant}
  \textsuperscript{\emph{e}} Ref.~\citenum{shui2020heterogeneous}
  \textsuperscript{\emph{f}} Ref.~\citenum{zhang2020molecular}
  \textsuperscript{\emph{g}} Ref.~\citenum{schutt2021equivariant}
  \textsuperscript{\emph{h}} Ref.~\citenum{liu2021transferable}
  \end{threeparttable}}}
% \end{sidewaystable}
\end{table*}

\subsection{Benchmark on extended systems}
Predicting properties of extended systems is a more ambitious task because of their intricate chemical environments. Since HermNet is capable to handle extended systems, we conduct this more challenging benchmark on the extended system datasets provided in Ref.~\citenum{zhang2018end}. The datasets contain properties of 16 different systems, in which Pt surface, Pt clusters on MoS$_2$ surface, and high entropy alloy (HEA) are three most difficult tasks. The Pt surface dataset includes data for Pt (100), (110) and (111) surfaces with different sizes and cells. The cluster-on-surface dataset includes 19 different kinds of Pt cluster on a MoS$_2$ slab. It is unfortunate that training on this dataset required too much computational time, so we chose not to further pursue this benchmark after some preliminary tuning (and no corresponding results are shown). The HEA dataset is explicitly divided into two datasets, such that the model should be trained on the first dataset which includes 40 kinds of 5 equi-molar-element CoCrFeMnNi HEA with random occupations and then tested on the test set in the first dataset and the entire second dataset that includes another 16 kinds of HEA with random occupations. Table~\ref{tab:3} shows the comparisons of root mean square errors (RMSEs) between DeepPot-SE/DeePMD~\cite{zhang2018end} and HVNet. Since the potential energy is an extended quantity, the RMSEs of energies were normalized with the system size in consistency with how the DeepPot-SE and DeePMD~\cite{zhang2018end} presented the results. As shown in Table~\ref{tab:3}, HVNet achieved lower RMSEs than DeepPot-SE on all tasks except Pt cluster on MoS$_2$ dataset, which we chose not to do due to the excessive amount of required training time. Detail of additional settings and specific discussions are provided in the Supplemental Material (SM)~\cite{SI}.

\begin{table*}
\centering
  {\caption{Comparison of the root mean square errors between DeepPot-SE (DeePMD) and HVNet trained on extended systems dataset, where the root mean square errors of the energies are normalized by the number of atoms in the system (energies in meV and forces in meV/\AA). The values in bold represent outperformance on the same task.}
  \label{tab:3}
  \resizebox{\textwidth}{43mm}{
  \vspace{2mm}
  \begin{threeparttable}
  \begin{tabular}{c|c|rrrrrrrr}
      \hline\hline
      \multirow{2}*{System} & \multirow{2}*{Sub-system} & \multicolumn{2}{c}{\textbf{DeepPot-SE\textsuperscript{\emph{a}}}}& ~ & \multicolumn{2}{c}{\textbf{DeepPMD\textsuperscript{\emph{a}}}} & ~ & \multicolumn{2}{c}{\textbf{HVNet}}\\
      \cline{3-10}
      ~ & ~ & energy & force & ~ & energy & force & ~ &  energy & force\\
      \hline
      bulk Cu & FCC solid & 0.18 & 90 & ~ & 0.25 & 90 & ~ & \textbf{0.107} & \textbf{84.97}\\
      \hline
      bulk Ge & Diamond solid & 0.35 & 38 & ~ & 0.60 & 35 & ~ & \textbf{0.283} & \textbf{22.04}\\
      \hline
      bulk Si & Diamond solid & 0.24 & 36 & ~ & 0.51 & 31 & ~ & \textbf{0.142} & \textbf{20.86}\\
      \hline
      bulk Al$_2$O$_3$ & Trigonal solid & 0.23 & 49 & ~ & 0.48 & 55 & ~ & \textbf{0.124} & \textbf{33.99}\\ 
      \hline
      \multirow{2}{*}{bulk C$_5$H$_5$N} & Pyridine-I & 0.38 & 25 & ~ & 0.25 & 25 & ~ & \textbf{0.060} & \textbf{17.78}\\
      \cline{2-10}
      ~ & Pyridine-II & 0.65 & 39 & ~ & 0.43 & 39 & ~ & \textbf{0.125} & \textbf{25.95}\\
      \hline
      \multirow{3}{*}{bulk TiO$_2$} & Rutile & 0.96 & 137 & ~ & 1.97 & 163 & ~ & \textbf{0.153} & \textbf{78.20}\\
      \cline{2-10}
      ~ & Anatase & 1.78 & 181 & ~ & 3.37 & 216 & ~ & \textbf{0.317} & \textbf{121.75}\\
      \cline{2-10}
      ~ & Brookite & 0.59 & 94 & ~ & 1.97 & 109 & ~ & \textbf{0.144} & \textbf{51.95}\\
      \hline
      \multirow{5}{*}{MoS$_2$+Pt} & MoS$_2$ slab & 5.26 & 23 & ~ & 17.2 & 34 & ~ &  \textbf{0.414} & \textbf{15.65}\\
      \cline{2-10}
      ~ & bulk Pt & 2.00 & 84 & ~ & 1.85 & 226 & ~ & \textbf{0.101} & \textbf{44.42}\\
      \cline{2-10}
      ~ & Pt surface & 6.77 & 105 & ~ & 7.12 & 187 & ~ & \textbf{0.90} & \textbf{85.99}\\
      \cline{2-10}
      ~ & Pt cluster & 30.6 & 201 & ~ & 25.4 & 255 & ~ & \textbf{1.853} & \textbf{43.45}\\
      \cline{2-10}
      ~ & Pt on MoS$_2$ & \textbf{2.62} & \textbf{94} & ~ & 5.89 & 127 & ~ & - & -\\
      \hline
      \multirow{2}*{CoCrFeMnNi HEA} & rand. occ. I & 1.68 & 394 & ~ & 6.99 & 481 & ~ & \textbf{0.342} & \textbf{304.02}\\
      \cline{2-10}
      ~ & rand. occ. II & 5.29 & 410 & ~ & 21.7 & 576 & ~ & \textbf{0.381} & \textbf{337.79}\\
      \hline\hline
  \end{tabular}
  \textsuperscript{\emph{a}} Ref.~\citenum{zhang2018end}
  \end{threeparttable}}}
\end{table*}

\section{Discussion}
\subsection{Model complexity}
The complexity of a sub-network is generally scaled as $\mathcal{O}(\vert \mathcal{N} \vert)$, where $\vert \mathcal{N} \vert$ is typically the number of the neighbors captured within a cutoff radius. The numbers of sub-networks for HVNet, HPNet and HTNet are $\mathcal{O}(N_e)$, $\mathcal{O}(N_e^2)$ and $\mathcal{O}(N_e^3)$, respectively. Here, $N_e$ is the number of element types present in the system. Therefore, HVNet is most useful when the number of distinct elements is large. Further discussions on the complexity analysis are deferred to the Supplemental Material (SM)~\cite{SI}.

\subsection{Perspective from the density functional theory}
To construct accurate force field for classical molecular dynamics simulations, potential energy surface needs to be reproduced up to first-principles precision. Actually, potential energy has hierarchical structure and can be decomposed into several terms as follows, 
\begin{equation}
    U = \sum_i E_i + \sum_{i\textless j} E_{ij} + \sum_{i\textless j \textless k} E_{ijk} + \cdots, \label{eqn:5}
\end{equation}
where the first term represents the energy of a single atom and the second term is the summation of all the pairwise interactions, such as the energy contributed from bonds. The third term denotes the three-body interactions, which typically entails angular specifications. Higher-order many-body interactions can be further included in order to build a more accurate potential energy surface. The layers shown in Fig.~\ref{fig:3} (b) and (c), which are equivalent to the message layer in the original PAINN proposal~\cite{schutt2021equivariant}, could be viewed as a single message passing neural network layer which models two body interactions since they merely process radial information. The inner products of the positional vectors presented in the modules in Fig.~\ref{fig:3} (d) and (e) or (f) are responsible for modeling three-body interactions. Thus the sub-network, i.e. concatenation of these layers, as shown in Fig.~\ref{fig:2} (a) and Fig.~\ref{fig:3} (a), exactly conforms to this hierarchical rule in Eq.~\ref{eqn:5}. 

On the other hand, graphs are constructed with a specific cutoff radius and only information of 1-hop neighbors is aggregated in a single MPNN layer. The final energy prediction is obtained with a global pool operation on all local environments. This suggests that locality is an essential property that facilitates the learning of potential energies. The DFT total energy could be expressed as a summation of eigenvalues of electronic Hamiltonian and the interaction of the nuclei with a correction to avoid double counting~\cite{martin2020electronic}. To take advantages of a localized basis as in a graph, we will discuss the total energy within the tight-binding framework, which could provide more physical insights. When the density is expressed as the superposition of spherical atomic densities~\cite{foulkes1989tight}, the total energy in the tight-binding representation is written as
\begin{align}
\begin{aligned}
    E_{\text{total}}
    &=\sum_{m, m^\prime}\rho_{m, m^\prime}H_{m, m^\prime} + \sum_{I<J}f(\vert \boldsymbol{R}_I - \boldsymbol{R}_J \vert), \label{eqn:5}
\end{aligned}
\end{align}
where $\rho_{m, m^\prime}$ is the density matrix. $H_{m, m^\prime}$ is the matrix element of the Hamiltonian between states $m$ and $m^\prime$, where $m = 1, \cdots, N_{\text{basis}}$ denote the states in the basis. $\boldsymbol{R}_I$ is the position of atom $I$, and $J$ is a neighboring site of $I$. The formula demonstrates that total energy could be decomposed into pairwise contributions, which is consistent with the layer made up of radial message passing layer in Fig.~\ref{fig:2} (a). Generally, the terms in Eq.~\ref{eqn:5} are both short-range interactions~\cite{kohn1996density, prodan2005nearsightedness, li2021deep} and could be extended to higher order interaction. Then the total energy could be expressed as $E_{\text{total}} = \sum_{i=I}^N \varepsilon_I^\prime$, which is a summation of local contributions from central particles. This indicates the locality of a system's overall energy, consistent with the idea underlying the seminal work of Ref.~\citenum{behler2007generalized}, which is widely adopted in the many follow-up works in this field.

\subsection{Heterogeneity in other frameworks}
In principle, the parameters of sub-networks in DeePMD~\cite{zhang2018end} are not shared for different element types, which is similar to heterogeneous graph neural networks. Thus the outperformance on extended systems results from the ability the sub-networks we used in this work. There are also other existing heterogeneous GNN frame work designed for MD simulations, but the design principle is very different. MXMNet~\cite{zhang2020molecular} utilized multiplex graphs, which could be viewed as heterogeneous graphs with individual node and two edge types, to capture global and local geometric information from multiplex graphs allocated with different cutoff radii. Heterogeneous molecular graph neural networks~\cite{shui2020heterogeneous} introduced heterogeneous graphs for molecules via grouping the original graph and a line graph into a single heterogeneous graph with two kinds of nodes. It processes information of nodes in original graph and line graph with two different graph neural networks respectively. The heterogeneity in these two works is equivalent to distinguishing original graphs and line graphs, which still treats the original graphs as a homogeneous graph.

\section{Summary and prospect}
In conclusion, we develop HermNet, a framework based on heterogeneous graph neural network, to learn multiple kinds of force fields in a single molecular dynamics simulation via extracting required subgraphs. Different from previous works, HermNet introduce heterogeneous graphs to describe different interactions of element types rather than to distinguish the hierarchy of the interactions. Among three variants of HermNet, we tested HVNet on a variety of systems, covering both molecular and extended systems, and obtained satisfactory results. Some discussions based on quantum mechanics and density functional theory have been provided to justify our model designs. Although we primarily focus on experiments with HVNet, in principle, HTNet is capable of modeling sophisticated interactions once enough data is provided. HVNet outperforms the state-of-the-art benchmark models on most of the tasks for small molecules. For the experiments on extended systems, HVNet also outperforms DeePMD~\cite{zhang2018end}. These results demonstrate the powerful representation and promising application potential of HVNet for diverse and intricate systems such as HEA. Finally, we emphasize that HermNet is a universal framework, whose sub-networks could be replaced by other advanced or specialized models. For example, unitary N-body tensor equivariant neural network (UNiTE)~\cite{qiao2021unite}, another remarkable framework based on the elegant group theory, was proposed recently, which performed impressively on molecular datasets. We believe that HermNet can deliver improved results by replacing the current sub-networks with UNiTE~\cite{qiao2021unite}. Besides, many-body interactions could also be truncated to higher order in sub-networks of HermNet, such as dihedral angular information~\cite{klicpera2021gemnet}. HermNet can also be extended to model interactions from higher order contributions via extracting higher-order subgraphs and invoking frameworks that model higher-order contributions properly.

\section{Methods}
HermNet is implemented with PyTorch~\cite{NEURIPS2019_9015} and Deep Graph Library~\cite{wang2019dgl} python library. Neighbors of the central particle are found by Scikit-Learn~\cite{scikit-learn} library and the node features are extracted by Atomic Simulation Environment~\cite{ase-paper} and Pymatgen~\cite{ong2013python} library. In our work, a simplified PAINN~\cite{schutt2021equivariant} is implemented as sub-network in both HVNet and HTNet. The angular formula in HVHet is the same as that in PAINN~\cite{schutt2021equivariant}, while that in HTNet is a little different. The proof that angular information could be introduced in HTNet with PAINN naturally is provided in Supplemental Material (SM)~\cite{SI}.

\section{Data availability}
The raw data of revised MD17, QM9, and bulk systems are available at \url{https://figshare.com/articles/dataset/Revised_MD17_dataset_rMD17_/12672038},  \url{https://deepchemdata.s3-us-west-1.amazonaws.com/datasets/molnet_publish/qm9.zip}, and \url{http://www.deepmd.org/database/deeppot-se-data/}, respectively.

\section{Code availability}
The implementations of HermNet described in the paper will be open source after the manuscript is accepted.

\section{Acknowledgments}
This work was supported by the Basic Science Center Project of NSFC (Grant No. 51788104), the Ministry of Science and Technology of China (Grants No. 2018YFA0307100 and 2018YFA0305603), the National Science Fund for Distinguished Young Scholars (Grant No. 12025405), the National Natural Science Foundation of China (Grant No. 11874035), Tsinghua University Initiative Scientific Research Program, and the Beijing Advanced Innovation Center for Future Chip (ICFC). The authors thank Tencent Quantum Lab for providing computational resources via Tencent Elastic First-principle Simulations (TEFS).

\section*{Competing interests}
The authors declare no competing financial interests.

% \begin{suppinfo}

% The Supporting Information is available free of charge at ...
% \begin{itemize}
%   \item Filename: brief description
% \end{itemize}

% \end{suppinfo}

\bibliography{ref}

\end{document}